\documentclass[letterpaper]{article} 
\usepackage{aaai24}  
\usepackage{times}  
\usepackage{helvet}  
\usepackage{courier}  
\usepackage[hyphens]{url}  
\usepackage{graphicx} 
\usepackage{natbib}  
\usepackage{caption} 
\usepackage{algorithm}
\usepackage{algorithmic}

\usepackage{xspace}
\usepackage{xcolor}

\usepackage{subcaption}

\usepackage{multirow}
\usepackage{pgfplots}
\pgfplotsset{compat=1.16}

\usepackage{newfloat}
\usepackage{listings}
\usepackage{amsfonts}
\usepackage{booktabs}
\usepackage{romannum}
\DeclareCaptionStyle{ruled}{labelfont=normalfont,labelsep=colon,strut=off} 
\floatstyle{ruled}
\newfloat{listing}{tb}{lst}{}
\floatname{listing}{Listing}
\title{Hyperparameter Optimization for Large Language Model Instruction-Tuning}
\author{
    Christophe Tribes\textsuperscript{\rm 1}, 
    Sacha Benarroch-Lelong\textsuperscript{\rm 1}, 
    Peng Lu\textsuperscript{\rm 2,3}, 
    Ivan Kobyzev\textsuperscript{\rm 2}
}
\affiliations{
    \textsuperscript{\rm 1}{GERAD and Polytechnique Montr\'eal}\\
    \{sacha.benarroch, christophe.tribes\}@polymtl.ca, \\
    \textsuperscript{\rm 2}{Huawei Noah’s Ark Lab, Montr\'eal} \\
    \textsuperscript{\rm 3}{RALI, Universit\'e de Montr\'eal}

}

\def\R{{\mathbb{R}}}

\newcommand{\nomad}{{\sf NOMAD}\xspace}

\newcommand{\mads}{{{\sc Mads}\xspace}}
\newcommand{\dmultimads}{{\sc DMulti-Mads}\xspace}

\begin{document}

\maketitle

\begin{abstract}
    The fine-tuning of Large Language Models (LLMs) has enabled them to recently achieve milestones in natural language processing applications. The emergence of ever larger LLMs has paved the way for more efficient fine-tuning methods. Among these, the Low-Rank Adaptation (LoRA) method keeps most of the weights of the pre-trained LLM frozen while introducing a low-rank decomposition of the weight matrix, enabling the tuning of only a very small proportion of the network. The performance on downstream tasks of models fine-tuned with LoRA heavily relies on a set of hyperparameters including the rank of the decomposition. In this work, we investigate the choice of these hyperparameters through two main blackbox optimization (BBO) techniques. We examine the whole pipeline of performing fine-tuning and validation on a pre-trained LLM as a blackbox and efficiently explore the space of hyperparameters with the \nomad algorithm, achieving a boost in performance and human alignment of the tuned model.


\end{abstract}

\section{Introduction}

Large-scale Language Models (LLMs) have shown exceptional ability in language understanding and generation~\cite{OPT, text2text, LLM_MTL, gpt3}. State-of-the-art models like ChatGPT~\cite{chatgpt} and GPT-4~\cite{gpt4} have garnered a great deal of interest from the academic and industrial communities. One of the main challenges of LLMs is how to control their behavior and make them follow specific instructions given by users~\cite{instruct_gpt}. Additional fine-tuning of LLMs on a dataset of instructions is called Instruction-Tuning; 
this technique has become ubiquitous due to its efficiency~\cite{zhang2023instruction_survey}. However, tuning large models demands a large amount of computer power. To overcome this, a common practice is to use Parameter Efficient Fine Tuning (PEFT) methods, which modify a limited selection of parameters in a pre-trained LLM while leaving the rest unchanged~\cite{peft}. Such methods are quite sensitive to the choice of hyperparameters~\cite{hu2021lora, Valipour2022DyLoRAPT}. In this work we investigate how hyperparameter optimization can better the instruct-tuning results. 

Hyperparameters selection by a human in order to tune a model is a tedious task but it can significantly improve model performance. Bergstra et al.~\citeyear{NIPS2011_86e8f7ab} suggest that hyperparameters optimization (HPO) forms the outer loop of a learning process. Applying an algorithmic approach to automate the process in finding better hyperparameters should also bring some efficiency. A grid search algorithm is a systematic but inefficient approach that tries a finite number of hyperparameters combinations. A blackbox optimization (BBO) algorithm should be a better choice for solving HPO efficiently within a fixed computational budget. 

In this work we 
investigated how two BBO solvers implementing different types of algorithms, namely \mads\xspace (a direct search algorithm implemented in \nomad) and TPE (a Bayesian model-based optimization algorithm implemented in NNI) 
behave when used to solve HPO for the Instruction-tuning of a specific LLM. We found different patterns in hyperparameter selection for these two 
optimizers, and assessed their effects on downstream tasks. Overall, we confirmed the necessity of careful HP selection in Instruction-tuning for performance boosting, both in downstream tasks and human preference. 



\section{Instruction-tuning Large Language Model}


Instruction-tuning has emerged recently as an important training paradigm~\cite{sanh_multitask_2022, wei_finetuned_2022, instruct_gpt, wang_self-instruct_2022}
to better adapt pre-trained models for human needs and enhance their ability to comprehend and respond to a diverse range of human requests.
Instruction-tuning is an additional training step for LLMs when the models are  fine-tuned on a dataset of instruction and
output pairs~\cite{alpaca, DatabricksBlog2023DollyV2, openassistant, FLAN}. It aims to bridge the gap between the next-word prediction objective of a language model and the users’ objective of having LLMs follow their instructions across various tasks and domains.


\subsection{Parameter-Efficient Fine-Tuning (PEFT)}
The success of Instruction-tuning heavily relies on a powerful model with at least several billion parameters.
Tuning of such models is usually difficult due to high computational costs in both time and memory. To circumvent this bottleneck, researchers developed Parameter-Efficient Fine-Tuning (PEFT) methods~\cite{peft}: instead of training all parameters, one freezes the majority of parameters in pre-trained models and only updates an incremental number of parameters. 

There are  various PEFT techniques generally falling into two groups:  Prompt Tuning~\cite{p_tuning_v2} when a few trainable tokens are added to the prompt; and different kinds of Adaptors~\cite{adaptor, He2021TowardsAU} when  extra trainable layers are inserted between layers of the pre-trained model.
In this work, we utilize the Low-Rank Adaptation (LoRA)~\cite{LoRA} method which adds trainable low-rank matrices to every model weights during training and merges the added parameters to the original pre-trained matrices for inference. The performance of LoRA-tuned models is very sensitive to the rank selection~\cite{Valipour2022DyLoRAPT}, hence the rank needs to be carefully picked for each dataset: too large rank could result in more overfitting on small datasets, yet a small rank may fail to capture the diversity of complicated instructions. Another important hyperparameter for LoRA is a scaling factor (LoRA-$\alpha$), which determines the scaling of low rank blocks that are added to the frozen parameters. We perform hyperparameters optimization (HPO) to select the optimal combination of these and some other LoRA hyperparameters to improve the performance of the tuned model.

\section{Hyperparameters Optimization}
In this work, the aim of hyperparameters optimization (HPO) is to obtain a fine-tuned model with the best performance measure. \nomad and NNI-TPE are considered for solving this HPO problem. 

\subsection{The \mads \xspace algorithm and \nomad}

\nomad\footnote{Available at
\url{https://www.gerad.ca/nomad}
and
\url{https://github.com/bbopt/nomad}.}~\cite{nomad4paper} is a software package for solving blackbox optimization (BBO) problems~\cite{AuHa2017} in which there is no analytical expressions for objective and constraint functions. 
The optimization problems have the following general form
	\begin{equation}
             \min_{x \in \mathcal X  \subseteq \R^n} \left\{  f(x) ~:~  c(x) \leq 0 \right\},
       \label{pb-genctr}
    \end{equation}
where $f: \mathcal X \subseteq \R^n \rightarrow  \R \cup \{\infty \}$ and
      $c: \mathcal X \subseteq \R^n \rightarrow (\R \cup \{\infty \})^m$
      are the given functions. 
      The function properties are not known and their evaluations are typically obtained after a computer program execution, with provided inputs and observed outputs. In addition, a blackbox function evaluation may take a significant amount of time and may fail to return valid outputs.
A HPO problem can be framed as a BBO problem where the objective function is linked to a performance measure of a model and the hyperparameters are the variables $x$.

\nomad implements the mesh adaptive direct search (\mads) algorithm~\cite{AuDe2006}. \mads\xspace is supported by a rigorous hierarchical convergence analysis based on various degrees of smoothness of the functions defining the problem. 
%
The \mads\xspace algorithm iterates {\em search} and {\em poll} steps to generate trial points on a {\em mesh} discretizing the space of variables. The search step generates trial points disseminated more globally in the space of variables. 
The poll step 
generates trial points around the current best solutions following rigid rules to ensure convergence to points satisfying some necessary optimality conditions. 
The mesh size may be adapted at each iteration. In addition, the mesh properties support by construction real variables, binary variables and granular variables
~\cite{AuLeDTr2018}. The mesh adaptation combined with the poll and search steps allows to explore more globally early during the optimization and more locally when the mesh is refined. This is one advantage of the \mads\xspace algorithm .

The \mads\xspace algorithm can handle general inequality constraints using the progressive barrier
~\cite{AuDe09a} approach to exploit the measure of constraint violation. 
\nomad includes BBO algorithms other than \mads. 
In particular, \dmultimads~\cite{BiLedSa2020} solves multiobjective optimization problems 
seeking detailed Pareto fronts. 
Hence, \nomad is suited to solve HPO problems with or without inequality constraints or that can have multiple objectives. 

\subsection{Neural Network Intelligence (NNI) toolkit}

Microsoft Neural Network Intelligence\footnote{Software available at \url{https://github.com/microsoft/nni}} (NNI) is an open-source toolkit to automate machine learning techniques such as hyperparameters optimization, model pruning, quantization, neural architecture search (NAS) and feature engineering.
Among the tuning algorithms available in NNI we have selected the Tree-structured Parzen Estimator~\cite{NIPS2011_86e8f7ab} (TPE) which is a Bayesian model-based optimization method. 
Bayesian optimization methods are appropriate to balance exploration and exploitation of the variable space with a limited evaluation budget.

TPE performs a series of optimization on a model of the objective function $f$  that is cheaper to evaluate (inner loop). A Gaussian Process (GP) is used to build the model and the inner loop aims to maximize the expect improvement (EI) of $f$. 
As new trial points are evaluated new models are fit based on the overall observation history. This process of sequential model-based optimizations~\cite{hutter2011sequential} (SMBO) can be repeated until the evaluation budget is used.
TPE is best suited for single objective HPO without inequality constraints.

\section{Experimental Setup}
\subsection{Instruction-tuning Settings}

\paragraph{Backbone Model}

LLaMA is a family of open-sourced large language models including models ranging from 7B to 65B parameters~\cite{touvron2023llama}. As our experiments aim at investigating the behavior of BBO algorithms, we conduct them with the 7 billions parameter version of LLaMA~2.\footnote{\url{https://huggingface.co/meta-llama/Llama-2-7b-hf}}
The fine-tuning of LLaMA~2 is done via the LoRA method. The method  has some specific hyperparameters that we explore with BBO (see Section BBO Settings below for details). 


\paragraph{Datasets} To perform our fine-tuning procedure, we use a mix of two same-structured instruction-following datasets (see Table~\ref{tab:training:eval:datasets} in Appendix). First is the 52k-entry dataset used in the Stanford Alpaca Project~\cite{alpaca}, that features a large diversity of instructions. 
Second is Databricks' Dolly dataset~\cite{DatabricksBlog2023DollyV2} containing 15k entries. We build a 54k-sized training set and a 13k-sized validation set, both containing 70\% of data from the Alpaca dataset and 30\% from Dolly, ensuring an identical distribution.




\paragraph{Training Details} The fine-tuning procedure minimizes the training loss by adapting LoRA trainable parameters. Once fine-tuned the validation loss of the model is computed. The HuggingFace Tranformers API~\cite{wolf2020huggingfaces} is used for handling the model, its training and validation on datasets. The default AdamW optimizer~\cite{loshchilov2019decoupled} is selected for training with a batch size fixed to 4. This pipeline is run on four NVIDIA-A100 GPUs with 80 GB memory. 

\subsection{BBO Settings}
In addition to LoRA rank, LoRA scaling $\alpha$ and dropout rate, we also seek to optimize the learning rate  that impacts the reduction of training loss (see Table~\ref{tab:nomad:map} in Appendix).

For the problem at hand, we can consider different types of performance measure.
The model training procedure indirectly seeks a fine-tuned model with low validation loss. But, the validation datasets are relatively small and a model may not generalize well. Hence, other performance measures on various instruction-following benchmarks are necessary to assess models' downstream capability but it would be very time-consuming if done during optimization. 

Moreover, considering multiple measures could require to use a multiobjective BBO formulation which demands a larger evaluation budget to obtain a refined Pareto front. In this work, to control the HPO computation time we chose a fixed and relatively small evaluation budget. Also, we decided to test if the validation loss computed at the last epoch can be used as the BBO single objective function. For validation on downstream tasks, we need to perform post-optimization assessments on several candidates. 

\section{Experimental Results}

\subsection{First optimization round}
A first optimization using \nomad was conducted to validate several a priori choices.
We started with a budget of 50 evaluations with 3 epochs. The validation loss is computed at each epoch. The duration of a single evaluation is around 2 hours and 15 minutes. It took less than 5 days to complete this optimization. 

\begin{figure}[t!]
    \centering
    \begin{tikzpicture}
        \begin{axis}[
            scatter/classes={a={mark=o,draw=black}},
            width=7.5cm,
            height=4.5cm,
            at={(0,0)},
            grid=both,
            xlabel={\scriptsize Evaluation \#},
            ylabel={\scriptsize Validation loss},
            ytick={0.7,0.8,0.9,1.0,1.1,1.2,1.2,1.3}
        ]
            \addplot+[scatter src=explicit symbolic,only marks, color=blue!70!white, mark=triangle] table [x=evaluation, y=eval_loss, col sep=comma] {nomad2-eval.csv};
        \end{axis}
    \end{tikzpicture}
    \caption{Objective value history. First \nomad optimization with 50 evaluations and a 3 epochs fine-tuning.}
    \label{fig:nomad:history:1}
\end{figure}
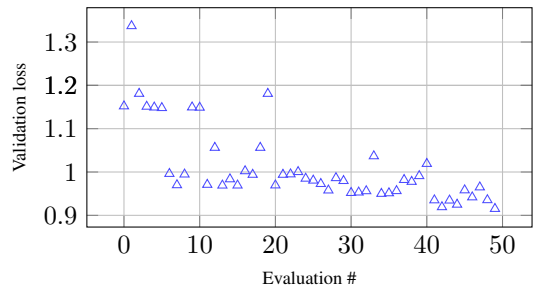

As expected, hyperparameter selection affects the fine-tuning training process. The smallest validation losses are obtained for evaluation points featuring the highest reduction in training losses. 
The best evaluation happens to be the last one. In addition, from the optimization history (see Figure~\ref{fig:nomad:history:1}) we can expect further reduction of the validation loss given an increased evaluation budget.
From the intermediate fine-tuning training steps (not shown here) we realize that on most evaluation points there is no significant change in validation loss between epoch 2 and epoch 3. Also, we observe that most of the best evaluation points have a LoRA rank value of $128$, which is the upper bound for this variable.

\subsection{Second optimization round}
For the next step, \nomad and NNI-TPE were used for HPO on 100 evaluations with 2 epochs. 
We also decided to increase the LoRA rank upper bound to $512$ in order to explore how it impacts model fine-tuning; in particular the capability to capture the diversity of instructions with the possible overfitting drawback.

The evaluation points obtained during the first round were given in a cache file to jump-start \nomad in the second round. These points were used during \mads\xspace search steps to construct quadratic models of the objective function and propose new promising trial points. 

\begin{figure}[t!]
    \begin{subfigure}{\linewidth}
        \centering
        \includegraphics[width=0.8\linewidth]{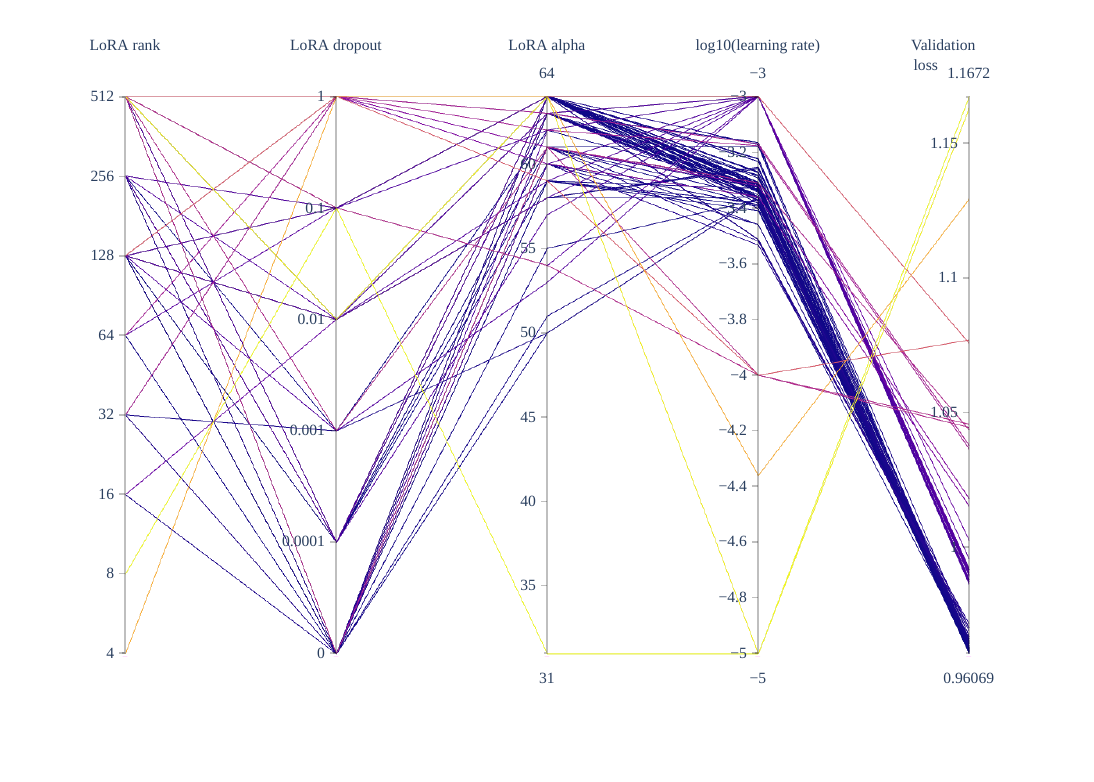}
        \caption{\nomad second round}
        \label{fig:parallel-plot-nomad3}
    \end{subfigure}
    \begin{subfigure}{\linewidth}
        \centering
        \includegraphics[width=0.8\linewidth]{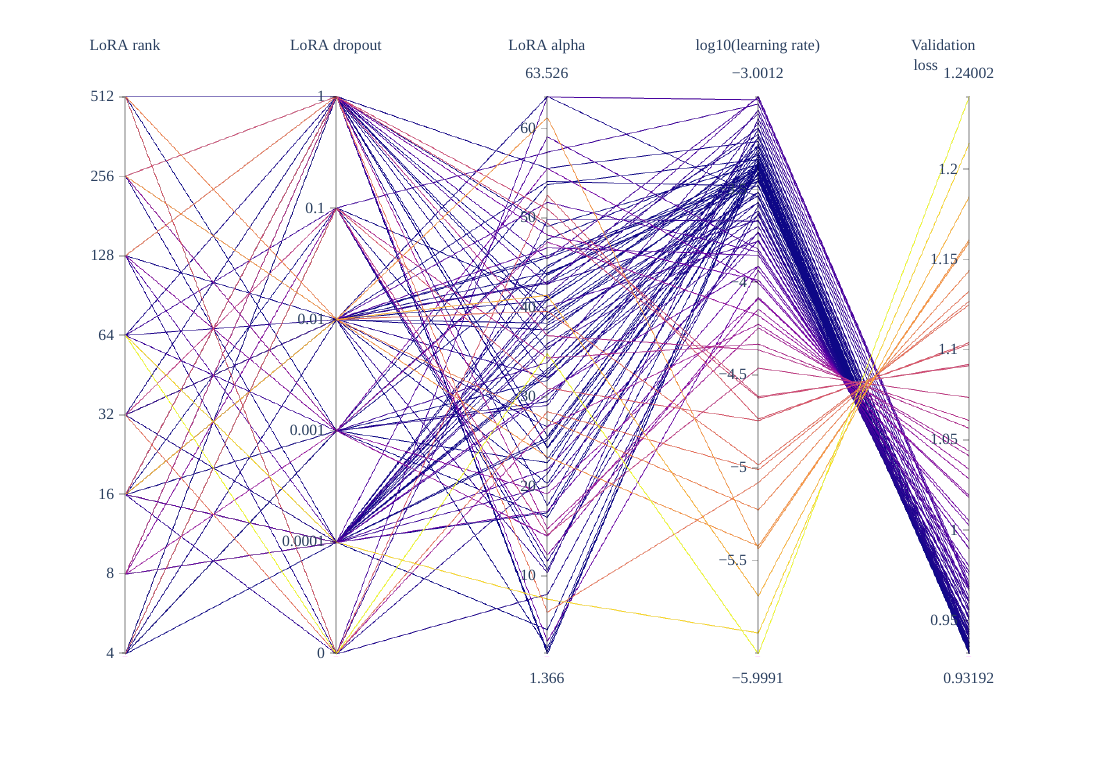}
        \caption{NNI-TPE}
        \label{fig:parallel-plot-nni}
    \end{subfigure}
    \caption{Parallel plots showing hyperparameters values and validation losses. Darker lines indicate lower validation losses.}
\end{figure}

\paragraph{\nomad results} Figure~\ref{fig:parallel-plot-nomad3} shows the hyperparameters combinations assessed by \nomad during this experiment and the validation loss yielded by the corresponding fine-tuned models. It makes clear that a learning rate around $10^{-3.5}$ and a scaling parameter $\alpha$ around $60$ yield the best results.  
Among the 10 best evaluations points, 5 have LoRA rank $r=512$ (including the best one), 4 have $r=256$ and 1 has $r=128$. The trend observed in first round 
linking large rank 
and lower validation loss is again observed. 
\nomad has obtained efficient hyperparameters combinations 
 in high-rank regions and has put the emphasis on exploitation through refining the other hyperparameters. By activating optional exploration methods in the search step, \nomad may have produced more trial points in low-rank regions.
Moreover, feeding the algorithm with a cache file from the first round may have introduced a bias in the search step in favor of these high-rank regions.

\paragraph{NNI-TPE results} Compared with the hyperparameters values tested by \nomad, NNI-TPE (see Figure~\ref{fig:parallel-plot-nni}) shows more variety confirming its explorative capability.
It also obtains evaluation points with lower validation losses compared to the best of \nomad.
Among the 10 best evaluations points of NNI-TPE, only 2 have a LoRA rank higher than $32$, the best one having rank $8$. This result shows that increasing LoRA rank is not the only way to obtain a lower validation loss, rather that low rank can perform well provided that other hyperparameters are chosen adequately.

\subsection{Evaluation of candidate best models}
Validation of the best candidate models was performed on downstream instruction-following tasks.
Instruct-Eval\footnote{\url{https://github.com/declare-lab/instruct-eval}}~\cite{chia2023instructeval} source codes and datasets are used to automate evaluation and obtain scores on a series of instruction-following tasks. The benchmarks considered in this work are MMLU~\cite{hendryckstest2021}, BBH~\cite{suzgun2022challenging}, DROP~\cite{Dua2019DROP} and HumanEval~\cite{chen2021evaluating} and are of quite different natures.


Table~\ref{tab:nomad:metrics} in Appendix shows the scores of the 10 best and 10 worst models (in terms of validation loss) explored by \nomad during the second optimization round.  In summary, the model ranked first does not give the best scores. Nevertheless, the 10 best models have very close validation losses. The 10 best models outscore the 10 worst models (including the one with default fine-tuning hyperparameters) and the baseline (without fine-tuning) for MMLU and HumanEval. 
For the BBH and DROP benchmarks the trend is not as clear.

Data from the 10 best models (validation loss) explored by \nomad (Table~\ref{tab:nomad:metrics}) and NNI-TPE (Table~\ref{tab:nni:scores}) optimizations is summarized in Table~\ref{tab:nni-tpe:nomad}. 
The 10 best MMLU, DROP and HumanEval scores are lower in average for NNI-TPE than what is obtained by \nomad even though NNI-TPE obtains the lowest validation loss. 
When judging by Instruct-Eval performance measures we can conclude that HPO using validation loss as objective function results in better models. However, lower validation losses do not necessarily translate into higher benchmark scores. 
With the current HPO problem formulation several candidates should be considered before selecting the best model for a downstream task.

\renewcommand\arraystretch{1.0}
\begin{table}[t!]
 \caption{Statistics of the 10 best models on downstream instruction-following tasks.}
    \centering
    \small
    \begin{tabular}{l l| c c c c c}\toprule
    \textbf{Method} & & \textbf{min} & \textbf{max} & \textbf{avg}. & \textbf{st}. \textbf{d}.\\
    \midrule
    \multirow{2}*{\nomad} & MMLU & 45.88 & 46.7 & 46.24 & 0.29\\
    & BBH & 32.07 & 32.99  & 32.50 & 0.25 \\
    & DROP & 29.67 & 30.95 & 30.28 & 0.45 \\
    & HumanEval & 14.63 & 18.9 & 16.94 & 1.52\\
    \midrule
    \multirow{2}*{NNI-TPE} & MMLU & 45.49 & 46.56 & 46.08 & 0.31\\
    & BBH & 32.27 & 34.43  & 32.93 & 0.42\\
    & DROP & 29.23 & 30.77 & 30.03 & 0.61 \\
    & HumanEval & 14.02 & 16.46 & 15.24 & 0.91\\
    \midrule
    \multirow{2}*{Default HPs} & MMLU & \multicolumn{4}{c}{43.56}\\
    & BBH &  \multicolumn{4}{c}{32.13} \\
    & DROP & \multicolumn{4}{c}{29.02} \\
    & HumanEval & \multicolumn{4}{c}{15.24}\\
    \bottomrule
    \end{tabular}
    \label{tab:nni-tpe:nomad}
\end{table}

\begin{figure}[t!]
    \centering
    \includegraphics[width=0.7\linewidth]{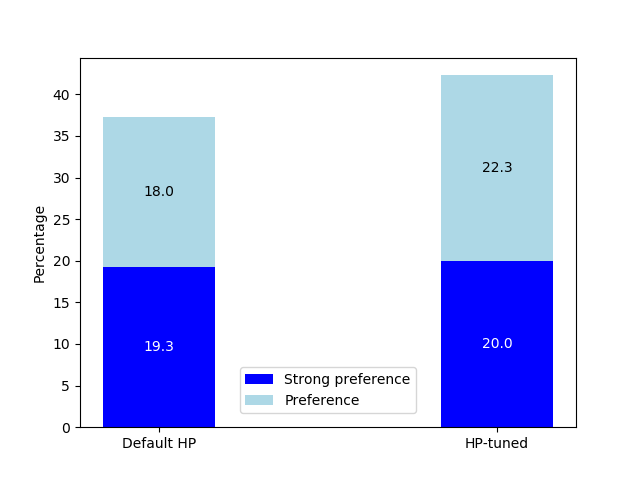}
    \caption{Human evaluation on the Vicuna human preference dataset.}
    \label{fig:human}
\end{figure}

\paragraph{Human Preference} We also conducted Human evaluation to check whether the generated results are aligned with human preferences. We sampled 30 questions randomly from the  Vicuna~\cite{vicuna2023} human preference dataset\footnote{\url{https://github.com/lm-sys/vicuna-blog-eval}} and asked Human evaluators to compare the answers generated by two models: the one tuned with \nomad as described above and the one with the default hyperparameters for LoRA. For each question, all evaluators are asked to judge which answer is better without knowing the source of answer. Figure~\ref{fig:human} shows that our HP-tuned model has a clear human preference compared to the default one by an overall preference score of 5\%. 

\section{Conclusion}

Hyperparameters optimization using blackbox optimization algorithms improves the performance of fine-tuned LLMs on downstream tasks and human evaluation. In particular, the best models are better than the model with default fine-tuning parameters. Also, for the 
three out of the four downstream tasks, the best candidate models are obtained by \nomad. 
NNI-TPE found candidate models with performance relatively close to those obtained by \nomad but with clearly lower LoRA ranks suggesting that different sets of hyperparameters may be optimal. More experiments should be conducted to either identify a single proper set of hyperparameters for LLM fine-tuning or to conclude that  hyperparameters optimization should form the outer loop for every LLM fine-tuning whenever possible.

The experiments show that validation losses are not perfectly aligned with downstream tasks scores. As future work we aim to develop an efficient and robust methodology to pickup a single best model. This can be achieved by guiding the blackbox optimization to consider more criteria into the HPO problem. Not all BBO algorithms offers enough flexibility to consider inequality constraints and multiple objectives. \nomad\xspace is a good option to handle such problems.


\section*{Acknowledgements}
This work is supported by the NSERC Alliance grant 544900-19 in collaboration with Huawei-Canada and by the NSERC Alliance-Mitacs Accelerate grant ALLRP~571311-21 (``Optimization of future energy systems'') in collaboration with Hydro-Qu\'ebec.

The authors want to thank Sébastien Le Digabel and Vahid Partovi Nia for their support and constructive comments.

\bibliography{bibliography}

\newpage
\appendix
\onecolumn
\section{Appendix: Datasets}

\begin{table}[H]
\caption{Instruction datasets used in this work. We report also the number of data samples and the average length of prompts (Avg.~L), the average length of completion (Avg.~C).}
\label{tab:training:eval:datasets}
\begin{center}\small
\begin{tabular}{l|rrrr}
\toprule
\textbf{Dataset} & \textbf{Type}                       & \textbf{\# samples} & \textbf{Avg.~L} & \textbf{Avg.~C }\\
\midrule
Alpaca  & LLM   & 52,002     & 27.8               & 64.6                   \\
Dolly   & Human & 15,011     & 118.1              & 91.3                  \\
\bottomrule
\end{tabular}
\end{center}
\end{table}

\section{Appendix: BBO settings}
Table~\ref{tab:nomad:map} gives the mapping between the variables handled by \nomad and the hyperparameters for evaluation. The type of variables, the bounds and initial values for the hyperparameters are also provided in this table. Contrary to \nomad, TPE does not require a special mapping between its variables and the hyperparameters; initial values are not required either.

Default values reported in Table are taken from the HuggingFace PEFT documentation.

\begin{table}[h]
\caption{Mapping \nomad variables into hyperparameters (right column) and initial values.}
    \centering
    \scriptsize
        \begin{tabular}{l|l|l}
        \toprule
           Rank &  Int. $r \in [1;8]$ &  $\rightarrow 2^{r+1} $ \\
           Dropout  & Int. $d \in [1;6]$ & $\rightarrow 0$ if $d=1$ \\
                    & &  else $10^{6-d}$ \\
           $\alpha$ &  Int. $\alpha \in [1;64]$  &  $\rightarrow \alpha $ \\
           LR  & Real $lr \in [-6; -3]$ & $\rightarrow 10^{lr}$ \\
        \midrule
           Default values & \multicolumn{2}{|l}{$r=2$ (Rank=8), $d=5$ (Dropout=0.1)} \\
            & \multicolumn{2}{|l}{$\alpha=32$, $lr=-5$ (LR=0.00001)} \\
        \bottomrule
    \end{tabular}
    
    \label{tab:nomad:map}
\end{table}

 \section{Appendix: Second round detailed results}
As our goal is a general-purposed model, we are also interested in Pareto optimality. A model is Pareto optimal if it is not dominated by any other model (among the ones evaluated). Picking-up a model is easier when the optimization returns a single Pareto optimal solution. Otherwise, Pareto optimal models have particular trade-offs between the different scores.


Table~\ref{tab:nomad:metrics} shows the scores of the 10 best and 10 worst models (in terms of validation loss) explored by \nomad during the second optimization round. We can note that no single model dominates all remaining models in Table~\ref{tab:nomad:metrics}. Interestingly, the models ranked 6 and 8 are Pareto optimal, whereas they do not achieve the best value for any score. In fact, a model outperforming in one kind of benchmark score can indicate its overspecialization. 

Table~\ref{tab:nni:scores} shows the scores of the 10 best and 10 worst models (validation loss) explored by NNI-TPE optimization. For BBH and DROP, when comparing \nomad and NNI-TPE, similar scores are obtained. The 10 best MMLU scores and HumanEval scores are lower for NNI-TPE than what is obtained by \nomad even though NNI-TPE obtains the lowest validation loss. 

\begin{table*}[h]
\caption{Instruct-Eval scores on the models generated by the two optimizers during the second optimization round. Models are ranked by increasing validation loss. The 10 best and 10 worst models are displayed. Best score values are in bold. $\star$ indicates Pareto optimality for this subset. $\dagger$ marks the model with default LoRA hyperparameters.}
    \centering
    \small
    \begin{subtable}[t]{0.45\textwidth}
        \centering
        \begin{tabular}{c c|c c c c}\toprule
            \multicolumn{2}{c |}{\textbf{Ranking}} & \textbf{MMLU} & \textbf{BBH} & \textbf{DROP} & \textbf{HumanEval}\\
            \multicolumn{2}{c |}{(valid. loss)} & & & &\\
            \midrule
            1 & & 45.94 & 32.51 & 29.71 & 17.07\\
            2 & $\star$ & 46.00 & 32.68 & \textbf{30.95} & 17.68\\
            3 & & 46.18 & 32.16 & 30.63 & 15.85\\
            4 & $\star$ & \textbf{46.70} & 32.37 & 30.15 & 18.29\\
            5 & & 46.42 & 32.07 & 30.33 & 18.29\\
            6 & $\star$ & 45.98 & 32.99 & 29.77 & 17.68\\
            7 & $\star$ & 46.46 & 32.50 & 30.95 & \textbf{18.90}\\
            8 & $\star$ & 46.57 & 32.60 & 29.67 & 14.63\\
            9 & & 46.28 & 32.42 & 30.29 & 16.46\\
            10 & & 45.88 & 32.67 & 30.39 & 14.63\\
            \midrule
            91 & & 42.48 & 31.43 & 28.62 & 12.20\\
            92 & & 42.47 & 32.30 & 28.40 & 12.80\\
            93 & & 42.44 & 30.45 & 28.62 & 12.80\\
            94 & $\star$ & 45.98 & \textbf{33.40} & 30.45 & 13.41\\
            95 & $\star$ & 45.09 & 32.77 & 30.85 & 15.24\\
            96 & & 42.32 & 30.98 & 29.01 & 13.41\\
            97 & &42.64 & 31.24 & 27.53 & 14.02\\
            98 & & 42.88 & 32.09 & 28.08 & 12.80\\
            99 & & 43.45 & 32.42 & 30.26 & 15.24\\
            100 & $\dagger$ & 43.56 & 32.13 & 29.02 & 15.24\\
            \midrule
             \multicolumn{2}{c |}{w/o fine-tuning} & 42.37 & 31.41 & 28.66 & 14.63\\
           
            \bottomrule
        \end{tabular}
        \caption{\nomad}
        \label{tab:nomad:metrics}
    \end{subtable}
    \hfill{}
    \begin{subtable}[t]{0.45\textwidth}
        \centering
        \begin{tabular}{c c|c c c c}
            \toprule
            \multicolumn{2}{c |}{\textbf{Ranking}} & \textbf{MMLU} & \textbf{BBH} & \textbf{DROP} & \textbf{HumanEval}\\
            \multicolumn{2}{c |}{(valid. loss)} & & & &\\
            \midrule
            1 & $\star$ & \textbf{46.56} & 32.41 & 30.26 & 14.63\\
            2 & $\star$ & 46.23 & \textbf{34.43} & 30.15 & 14.02\\
            3 & $\star$ & 46.28 & 32.86 & 29.28 & \textbf{16.46}\\
            4 & $\star$ & 46.40 & 32.27 & 29.77 & 15.85\\
            5 & $\star$ & 45.94 & 32.83 & 30.58 & 14.02\\
            6 & $\star$ & 45.84 & 33.49 & 30.25 & \textbf{16.46}\\
            7 & & 46.13 & 32.3 & 29.72 & 14.63\\
            8 & $\star$ & 46.06 & 32.9 & 30.34 & 15.85\\
            9 & $\star$ & 45.91 & 32.78 & 30.77 & 15.24\\
            10 & & 45.49 & 33.03 & 29.23 & 15.24\\
            \midrule
            91 & & 43.16 & 32.02 & 28.91 & 14.63\\
            92 & & 43.10 & 32.38 & 29.79 & 15.85\\
            93 & & 43.27 & 31.54 & 29.26 & 14.02\\
            94 & & 43.46 & 31.55 & 28.80 & 14.02\\
            95 & & 43.42 & 31.65 & 28.94 & 14.63\\
            96 & & 42.81 & 32.05 & 29.23 & 14.02\\
            97 & & 43.01 & 31.45 & 32.41 & 14.02\\
            98 & $\star$ & 42.86 & 32.00 & \textbf{32.94} & 14.63\\
            99 & $\star$ & 46.20 & 32.04 & 32.78 & 15.85\\
            100 & & 46.05 & 31.24 & 28.89 & 14.02\\
            \midrule
             \multicolumn{2}{c |}{w/o fine-tuning} & 42.37 & 31.41 & 28.66 & 14.63\\
            \bottomrule
        \end{tabular}
        \caption{NNI-TPE}
        \label{tab:nni:scores}
    \end{subtable}
    
\end{table*}

\section{Appendix: Human Evaluation Setting}
The evaluation is conducted with Google Forms
with 30 instructions in that form. The ordering of the questions and the responses are totally randomized. We found 10 experienced volunteering
annotators who are fluent in English and hold bachelor’s degrees or above. 
\end{document}